\documentclass[conference]{IEEEtran}
\IEEEoverridecommandlockouts

\IEEEoverridecommandlockouts

\usepackage{cite}
\usepackage{amsmath,amssymb,amsfonts}
\usepackage{algorithmic}
\usepackage{graphicx}
\usepackage{textcomp}
\usepackage{xcolor}
\def\BibTeX{{\rm B\kern-.05em{\sc i\kern-.025em b}\kern-.08em
    T\kern-.1667em\lower.7ex\hbox{E}\kern-.125emX}}

\usepackage{tikz}
\usetikzlibrary{shapes}
\usetikzlibrary{arrows}
\usetikzlibrary{automata}
\usetikzlibrary{fit}
\usetikzlibrary{shadows}
\usetikzlibrary{shapes}
\usetikzlibrary{trees}
\usetikzlibrary{positioning}
\usetikzlibrary{patterns}
\usetikzlibrary{backgrounds}
\usetikzlibrary{matrix}
\usetikzlibrary{pgfplots.groupplots}
\usetikzlibrary{external}

\usepackage{pgfplots}
\pgfplotsset{compat=1.18}

\tikzstyle{every state}=[draw=black,text=black,inner color= white,outer color= white,draw= black,text=black]
\tikzstyle{place}=[thick,draw=sthlmBlue,fill=blue!20,minimum size=6mm, opacity=.5]
\tikzstyle{red place}=[square,place, draw=sthlmRed, fill=sthlmLightRed]
\tikzstyle{green place}=[diamond, place, draw=sthlmGreen, fill=sthlmLightGreen]
\tikzset{chance/.style={state,place}}
\tikzset{maxnod/.style={state,red place}}
\tikzset{minnod/.style={state,green place}}
\tikzset{termin/.style={align=left}}

	\definecolor{sthlmLightBlue}{RGB}{214,237,252} 
		
	\definecolor{sthlmBlue}{RGB}{0,110,191} 
		
	\definecolor{sthlmLightGreen}{RGB}{213,247,244} 
		
	\definecolor{sthlmGreen}{RGB}{0,134,127} 

	\definecolor{sthlmLightGrey}{RGB}{213,217,225} 
		
	\definecolor{sthlmGrey}{RGB}{245,243,238} 
		
	\definecolor{sthlmDarkGrey}{RGB}{51,51,51} 

	\definecolor{sthlmLightOrange}{RGB}{255,215,210} 
		
	\definecolor{sthlmOrange}{RGB}{221,74,44} 

	\definecolor{sthlmLightPurple}{RGB}{241,230,252} 
		
	\definecolor{sthlmPurple}{RGB}{93,35,125} 

	\definecolor{sthlmLightRed}{RGB}{254,222,237} 
		
	\definecolor{sthlmRed}{RGB}{196,0,100} 
		
	\definecolor{sthlmYellow}{RGB}{252,191,10} 

\usepackage[framemethod=TikZ]{mdframed}

\newcommand{\textalgo}[1]{\mathrm{#1}}
\newcommand{\nbplayer}[0]{\mathcal{N}}

\usepackage{amsmath}
\usepackage{amsfonts}
\usepackage{mathrsfs}
\usepackage{amssymb}
\usepackage{amsthm}
\usepackage{mathabx}

\usepackage{xspace}
\def\ie{{\em i.e.}\xspace}
\def\eg{{\em e.g.}\xspace}

\newtheorem*{definition}{Definition}

\usepackage{thmtools}
\usepackage{thm-restate}
\usepackage{cleveref}

\usepackage{array,multirow,makecell}
\usepackage[mathcal]{euscript}
\usepackage{paralist}
\usepackage{cancel}
\usepackage{diagbox}
\usepackage{graphicx}
\usepackage{booktabs}       
\usepackage{rotating}
\usepackage{algorithm2e}
\usepackage{subfig}

\newcommand{\budget}{budget}
\newcommand{\move}{m}
\newcommand{\allMove}{Moves}
\newcommand{\PerfectAlgo}{PerfectAlgo}
\newcommand{\InfostateSampling}{InfoSampling}

\newcommand{\infostate}{s}
\newcommand{\worldstate}{w}
\newcommand{\treestate}{u}
\newcommand{\score}{score}
\SetKwProg{Fn}{Function}{:}{}
\RestyleAlgo{ruled}

\title{Mixture of Public and Private Distributions in Imperfect Information Games.}

\author{\IEEEauthorblockN{Jérôme Arjonilla}
\IEEEauthorblockA{\textit{Paris Dauphine University - PSL}\\
Paris, France \\
jerome.arjonilla@hotmail.fr}
\and
\IEEEauthorblockN{Abdallah Saffidine}
\IEEEauthorblockA{\textit{University of New South Wales}\\
Sydney, Australia \\
abdallah.saffidine@gmail.com}
\and
\IEEEauthorblockN{Tristan Cazenave}
\IEEEauthorblockA{\textit{Paris Dauphine University - PSL}\\
Paris, France \\
Tristan.Cazenave@dauphine.psl.eu}
}

\begin{document}

\maketitle
\IEEEpubidadjcol

\begin{abstract}

In imperfect information games (e.g. Bridge, Skat, Poker), one of the fundamental considerations is to infer the missing information while at the same time avoiding the disclosure of private information. Disregarding the issue of protecting private information can lead to a highly exploitable performance. Yet, excessive attention to it leads to hesitations that are no longer consistent with our private information. In our work, we show that to improve performance, one must choose whether to use a player's private information. We extend our work by proposing a new belief distribution depending on the amount of private and public information desired. We empirically demonstrate an increase in performance and, with the aim of further improving performance, the new distribution should be used according to the position in the game. Our experiments have been done on multiple benchmarks and in multiple determinization-based algorithms (PIMC and IS-MCTS).

\end{abstract}


\section{Introduction}

Search in artificial intelligence has been constantly evolving over the last few decades, and game-oriented research has always been a cornerstone of this success. Chess, Go \cite{Silver2016MasteringTG, silver_mastering_2017, silver_general_2018}, Poker \cite{brown_superhuman_2019}, Skat, Contract Bridge or Dota \cite{berner_dota_2019} are among the most famous ones. 

Perfect information games (Chess, Go) --- where all information is available for each player --- have been the most studied, and many algorithms have been able to achieve a level far beyond the level of a human professional player. On the other hand, Imperfect Information Games (IIGs) (Poker, Skat, Bridge) --- where some information is hidden --- have been less studied, and only a few algorithms are capable of beating professional human player \cite{tammelin_solving_2015,brown_superhuman_2019}. 

In IIG, the complexity is heightened by the missing information, as one must try to infer the missing information of the opponents and, at the same time, be wary to not reveal private hidden information to opponents. Among the methods used in IIG, determinization-based algorithms --- where the hidden information is fixed according to a belief distribution --- such as Perfect Information Monte Carlo (PIMC) \cite{long_understanding_2010}, Recursive PIMC \cite{furtak_recursive_2013}, Information Set MCTS \cite{cowling_information_2012} or AlphaMu \cite{cazenave_alpha_2021} achieve state-of-the-art performance in many trick-taking card games (Contract-Bridge, Skat).

In the work cited above, the determinization operates by sampling the hidden information according to the private information of a given player, \ie what has happened since the beginning, from the point of view of a given agent. However, by doing so, one can indirectly reveal private information to opponents, which can lead to a highly exploitable performance. 

Recently, the concept of public knowledge \cite{kovarik_rethinking_2022} --- where a distinction is made between observations accessible to everyone and those accessible individually --- has emerged in IIGs. This concept has resulted in many breakthroughs thanks to the decomposition, which made the calculations feasible \cite{moravcik_deepstack_2017,brown_combining_2020}. Despite this large benefit, there are limitations to its use, especially in the context of belief distribution. By doing so, we completely remove the knowledge observed by the acting player, and one might wonder whether not using the private information was useful.

In this work, we analyze the impact of using one method rather than another and present a new belief distribution, which is a mixture of both public and private belief distribution. We extend the study by analyzing different mixtures, depending on the position within the game. Our experiments are carried out on determinization-based algorithms, which use the belief distribution to fix the incertitude. 



The paper is organized as follows: Section \ref{sec:nota} presents notation and current determinization-based algorithms; Section \ref{sec:belief_distribution} explains the different belief distributions used with their advantages and drawbacks, and presents our new belief distribution; Section \ref{sec:exp} empirically shows that using the new belief distribution allows us to improve past performance; and the last section summarizes our work and future work.

\section{Notation and Background}
\label{sec:nota}
\subsection{Notation}

We use the notation based on factored-observation stochastic games (FOSGs \cite{kovarik_rethinking_2022}). This formalism distinguishes between private and public observations. 

A Game G is composed of the following elements. The set of agents $\nbplayer = \{1, 2, \dots, N\} $ agents, the set of \textbf{world state} possible $\mathcal{W}$. In each world state $w\in \mathcal{W}$, the acting player $i$ chooses an action $a \in \mathcal{A}(w)$, where $\mathcal{A}(w)$ denotes the legal actions at $w$. After an action $a$, we reach the next world state $w'$ from the probability distribution of playing $a$ in $w$. 

During the transition from $w$ to $w'$ by playing $a$, two observations are received: a \textbf{public observation} and a \textbf{private observation}. Public observation is the observation visible by every player noted $o_{pub} \in \mathcal{O}_{pub}(w,a,w')$ where $ \mathcal{O}_{pub}(w,a,w') $ refers to all the possible public observations. Private observation is the observation visible by a precise player $i$, noted $ o_{priv(i)} \in \mathcal{O}_{priv(i)}(w,a,w')$ where $ \mathcal{O}_{priv(i)}(w,a,w') $ refers to all the possible private observations.

A \emph{history} is a finite sequence of legal actions and world states, denoted $h^t = (w^0,a^0,w^1,a^1, ..., w^t)$. For describing the point of view of an agent $i$ of a history $h$, we introduce an \textbf{infostate} $s_i(h)$. An \textbf{infostate} for agent $i$ is a sequence of an agent’s observations and actions $s^t_i$ = ($o_i^0$, $a^0_i$, $o_i^1$, $a^1_i$, ..., $o_i^t$) where $o_i^k$ = ($o_{pub}^k$, $o^k_{priv(i)}) $. A \textbf{public infostate} is a sequence of public observations $s^t_{pub} = (o_{pub}^0, o_{pub}^1, ..., o_{pub}^t)$.

\textbf{Determinization} refers to the fact that we sample a world state according to a belief distribution of the world states possible. Determinizing the belief distribution is not new and a similar concept exists in other formalisms such as belief state in Partially Observable Markov Decision Process (POMDP) problems \cite{Smith-2007-9778} or occupancy-state in Decentralised-POMDPs problems \cite{dibangoye:hal-01279444}. 

\subsection{Determinization-based algorithms}
\label{sub:backgroup}

Each determinization-based algorithm has its own characteristics. Nevertheless, they share some common features such as (\romannumeral 1) sampling a world state according to a belief distribution over the possible world states; and (\romannumeral 2) using a perfect information algorithm for estimating the value of the sampled world state. 

The algorithms are simple and, in practice, they achieve great results, mainly due to the use of perfect information algorithms (AlphaBeta \cite{knuth_analysis_1975}, MCTS \cite{browne_survey_2012} or Value Network) that are fast and efficient.
\\\\
In the following, we present two determinization-based algorithms that are baseline and will, at a later stage, be used in our experiments.

\subsubsection{PIMC}

Perfect Information Monte Carlo (PIMC) is the state of the art of many IIG  problems such as Contract-Bridge, Skat, and many others. 

The algorithm is defined in Algorithm \ref{alg:pimc} and works as follows: (\romannumeral 1) samples a world state by using the player's private information; (\romannumeral 2) plays each action of the sampled world state; (\romannumeral 3) estimates the reward of the new world state by using an algorithm available in perfect information setting; (\romannumeral 4) repeats until the budget is over; and, (\romannumeral 5) selects the action that produces the best result in average. In practice, PIMC often uses AlphaBeta as the perfect information evaluator.

\begin{algorithm}
\caption{PIMC}\label{alg:pimc}
\SetKwFunction{FMain}{PIMC}
\SetKwProg{Fn}{Function}{:}{}
\Fn{\FMain{$\textalgo{\infostate} $}}{
    \For{$\textalgo{\move}$ $\in$ $\textalgo{\allMove}$ ($\textalgo{\infostate})$}{
        $\textalgo{\score}$[$\textalgo{\move}$] $\gets$ $0$\;
    }
    \While {$\textalgo{\budget}$} {
        $\textalgo{\worldstate}$ $\gets$ $\textalgo{\InfostateSampling}$($\textalgo{\infostate}$)\;
        
        \For{$\textalgo{\move}$ $\in$ $\textalgo{\allMove}$ ($\textalgo{\worldstate}$)}{
            $\textalgo{\score}$ [$\textalgo{\move}$] $\gets$ $\textalgo{\score}$[$\textalgo{\move}$] + $\textalgo{\PerfectAlgo}$ ($\textalgo{\worldstate}$, $\textalgo{\move}$)\;
        }
    }
    \Return{Best action on average}
}
\end{algorithm}

\subsubsection{IS-MCTS}

Information Set Monte Carlo Tree Search \cite{cowling_information_2012} uses Monte Carlo Tree Search (MCTS) \cite{browne_survey_2012} according to a sampled world state. 

MCTS is a state-of-the-art tree search algorithm in perfect information games. It works as follows (\romannumeral 1) selection --- selects a path of nodes based on an exploitation policy; (\romannumeral 2) expansion --- expands the tree by adding a new child node; (\romannumeral 3) playout --- estimates the child node by using an exploration policy; and, (\romannumeral 4) backpropagation --- backpropagates the result obtained from the playout through the nodes chosen during the selection phase. In practice, MCTS often uses random playout as the perfect information evaluator, and UCB1 in the selection phase.

IS-MCTS works by using MCTS according to a sampled world state, \ie the selection and playout are done on the sampled world state. 

\begin{algorithm}
\caption{IS-MCTS}\label{alg:is_mcts}
\SetKwFunction{FMain}{IS-MCTS}
\Fn{\FMain{$\textalgo{\infostate} $}}{

    \While {$\textalgo{\budget}$} {
        $\textalgo{\worldstate}$ $\gets$ $\textalgo{\InfostateSampling}$($\textalgo{\infostate}$)\;

        $\textalgo{MCTS}$ conditioned on $\textalgo{\worldstate}$.\;
    }
    \Return{Normalise visit count for each action}
}

\SetKwFunction{FMain}{MCTS}
\Fn{\FMain{$\textalgo{\worldstate} $}}{
        $\textalgo{\treestate}$ $\gets$ $\textalgo{Selection}$($\textalgo{\worldstate}$)\;
        $\textalgo{\treestate}$ $\gets$ $\textalgo{Expansion}$ ($\textalgo{\treestate}$,$\textalgo{\worldstate}$)\;
        $\textalgo{\treestate}$ $\gets$ $\textalgo{Simulation}$ ($\textalgo{\treestate}$,$\textalgo{\worldstate}$)\;
        $\textalgo{Backpropagation}$($\textalgo{\treestate})$\;
}
\end{algorithm}

\section{Belief Distributions}
\label{sec:belief_distribution}

To present the different belief distributions, with their advantages and drawbacks, we use the following example throughout the section to facilitate understanding. 

The example is based on the famous game `Liar's Dice' (an explanation of the game is given in Subsection \ref{subsub:liarDice}). In our case, two players, each with $1$ die of $2$ sides. We denote $\{P_1:X; P_2:Y\}$ for player $1$ has $X$ and player $2$ has $Y$. There are four world states possible ($w_1=\{P_1:1;P_2:1\}$, $w_2=\{P_1:1;P_2:2\}$; $w_3=\{P_1:2;P_2:2\}$, $w_4=\{P_1:2;P_2:1\}$). 

For each player, there are two infostates possible and one public infostate $s_{pub} = \{o^1_{pub}= \emptyset, o^2_{pub}=\emptyset\}$ (no observation). For the player $1$ we have $s_1=\{o^1_{priv(1)}= 1, o^2_{priv(1)}=\emptyset \}$ or $s'_1 =\{o^1_{priv(1)}= 2, o^2_{priv(1)}=\emptyset\}$ (\ie Player $1$ observes the die rolled but not the die rolled by the other player), and for the player $2$, we have  $s_2 =\{o^1_{priv(2)}= \emptyset, o^2_{priv(2)}=1\}$ or $s'_2 =\{o^1_{priv(2)}=\emptyset, o^2_{priv(2)}=2\}$ (\ie Player $2$ observes the die rolled but not the die rolled by the other player).

\begin{figure}[!htbp]
\centering

\includegraphics[scale=0.20]{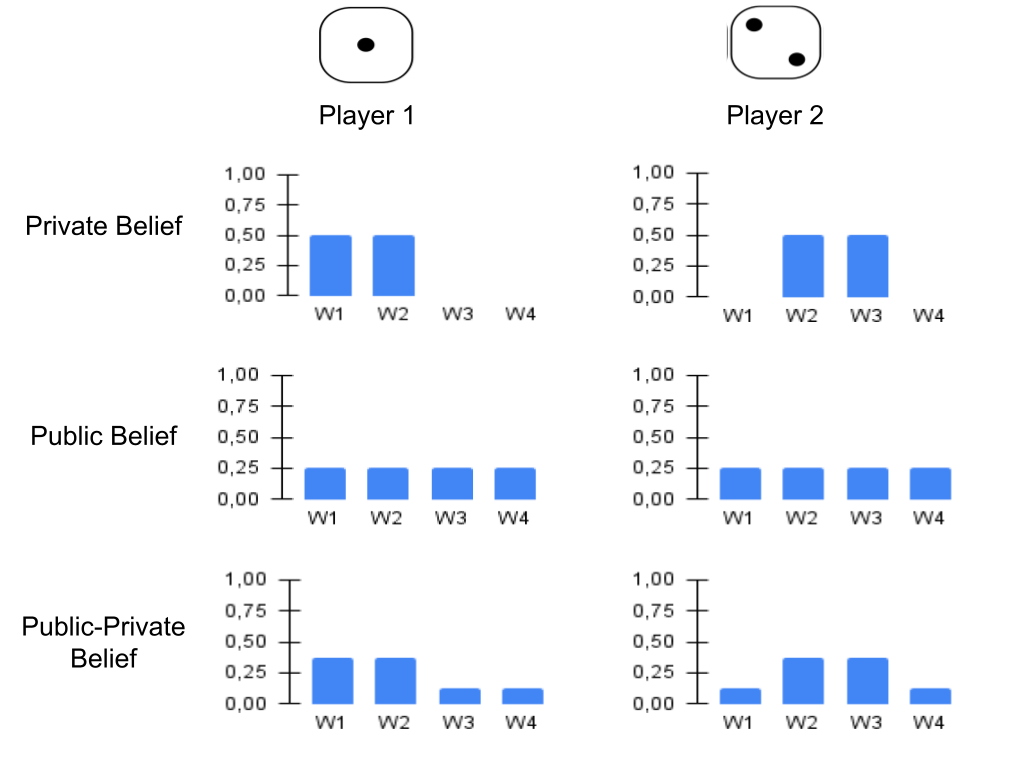}

\caption{Multiple belief distributions for the game Liar's Dice with $1$ dice of $2$ sides each. Four world states possible $w_1$, $w_2$, $w_3$ and $w_4$. The Public-Private belief uses the mixture distribution with $\lambda=0.5$.}

\label{fig:beliefdistribution}
\end{figure}

In the following, we suppose that the world state of this example is $w_2$. Therefore, for the player $1$, the infostate is $s_1$ with two world states possible ($\{w_1,w_2\}$) and for the player $2$, the infostate is $s'_2$ with two world states possible ($\{w_2;w_3\}$). Fig. \ref{fig:beliefdistribution} represents the different belief distributions presented throughout the section.

\subsection{Private Distribution}
\label{privateDistribution}

As previously introduced, current determinization-based algorithms work by sampling world states according to the player's private information distribution, \ie knowing a player's private and public observation, we sample a world state.

Let $S_j(s_i)$ be the set of possible infostates for player $j$ conditioned on the infostate $s_i$ of the player $i$. 
In our example, the infostate possible for the player $2$ when the player $1$ has $s_1$ is $S_2(s_1) = \{s_2;s'_2\}$, \ie having the die $1$ for the player $1$ does not exclude the player $2$ to have a 1 or a 2. Depending on the game $S_j(s_i)$ can be restrictive, \eg in trick-taking card games if the player $i$ has the card `Queen of Hearts', no opponent can have it. 

\begin{definition}[Private Belief Distribution]
Let $S_j(s_i)$ be the set of possible infostates for player $j$ conditioned on the infostate $s_i$. Let $\Delta S_j (s_{i})$ denotes the probability distribution over the elements of $S_j (s_i)$. We define the private belief distribution as $\Delta_{i}(s_i) = (\Delta S_1 (s_i), \dots, \Delta S_i(s_i) ,\dots, \Delta S_N (s_i)) = (\Delta S_1 (s_i), \dots, s_i , \dots, \Delta S_N (s_i))$ .
\end{definition}

In Fig. \ref{fig:beliefdistribution}, using Player 1's private belief state provides the following belief distribution $\Delta_1(s_1)= (\{s_1 : 100\%\}, \{s_2 : 50\%; s'_2 : 50\%\})$, which results in two equiprobable world states ($w_1$, $w_2$).
\\\\
When using the private distribution for determinization, the algorithm samples a world state ($w_1$ or $w_2$) consistent with the current player's information ($s_1$) and, as the state-of-the-art in trick-taking game shows, great performance is obtained. Yet, by doing so, 3 problems arise. 

(\romannumeral 1) It is not consistent with the other player's belief, \eg if we use it with the first player, the algorithm samples $w_1$ or $w_2$ but never $w_3$, which is nevertheless, a world state possible from the point of view of the player $2$.

(\romannumeral 2) It is not able to mislead others. In our example, two actions are possible for the first player, `I have a one' and `I have a two'. The action `I have a two' is a lie, however, one may want to play this action with the aim of deceiving the opponent. However, in our case only $w_1$ or $w_2$ can be sampled and, in each world, the action `I have a two' results in a defeat because the second player will say `This is a lie'. Therefore, lying is never an option, as it never succeeds. 

(\romannumeral 3) It, indirectly, allows the opponents to infer our private information, \eg after playing multiple matches, the second player understands that, if the first player plays `I have a two', it is because he really has a two as it can not lie, and therefore, play to counter it.

Trying to infer missing information is one of the key components of IIG, and using the private belief distribution could result in a highly exploitable performance. To remove this problem, one can use public belief distribution, as presented in the next section. 

\subsection{Public Distribution}
\label{publicDistribution}

Recently in IIG, many algorithms \cite{moravcik_deepstack_2017,brown_combining_2020} have been using the concept of public observation. This concept has resulted in many breakthroughs thanks to decomposition, which made the calculations feasible. One application of public observation is the creation of a public belief distribution over the world states possible according to the public observations observed so far.

\begin{definition}[Public Belief Distribution \cite{brown_combining_2020}]
Let $S_j(s_{pub})$ be the set of possible infostates for player $j$ conditioned on the public infostate $s_{pub}$. Let $\Delta S_j (s_{pub} )$ denote the probability distribution over the elements of $S_j (s_{pub})$. We define the public belief distribution as $\Delta_{pub} (s_{pub})= (\Delta S_1(s_{pub}), ..., \Delta S_N (s_{pub}))$.
\end{definition}

In our example, using the public belief distribution from the point of view of the player $1$ would result in the following belief distribution $\Delta_{pub} = (\{s_1 : 50\%; s'_1 : 50\%\}, \{s_2 : 50\%; s'_2 : 50\%\}$. In other world, every world state are equiprobable, this is due to the public infostate that does not contain any information. 
\\\\
Using a public belief distribution instead of a private belief distribution removes the problem defined in  Section \ref{privateDistribution}. 

(\romannumeral 1) It is consistent with the other player's doubts, \eg it samples the world $w_3$ which is a world state possible of the second player. 

(\romannumeral 2) It is capable of misleading others, \eg when sampling $w_3$ or $w_4$ the action `I have a two' does not result in a defeat for the first player, therefore, allows the first player to play the action `I have a two'. 

(\romannumeral 3) It no longer reveals private information, \ie as the reasoning is no longer biased toward the private information, it can not be used against it. 

Nevertheless, using public distribution has a significant drawback as it does not consider a player's private information, and one might wonder whether it is useful to not use private information. It is straightforward to consider that the extent to which private information should be kept hidden depends on the game being played and, in certain games, it is not necessary to keep the information concealed.

In addition, by using public distribution, one must be aware as there are more world states possible (\eg by using private distribution, we have two world states possible and by using public distribution, we have four world states possible), which can be intractable in large games.

\subsection{Mixture between public and private distribution}

To solve both of the problems defined in Section \ref{privateDistribution} and in Section \ref{publicDistribution}, we propose to use a mixture of private and public distribution. 

\begin{definition}[Mixture Belief Distribution]
Let $s_{pub}$ be the public infostate associated with the infostate $s_i$. 
We define the mixture belief distribution as $\Delta_{\lambda}(s_i) = (1-\lambda) \Delta_{i}(s_i) + \lambda \Delta_{pub}(s_{pub})$
\end{definition}

The mixture belief distribution allows us to be consistent with the problem encountered. When care must be taken not to reveal information, one can increase $\lambda$. In contrast, when it is not appropriate to withhold information, one can decrease $\lambda$. The private belief distribution is obtained when $\lambda = 0$ and the public belief distribution is obtained when $\lambda=1$.
\\\\
In our example, when using the mixture with $\lambda=0.5$ for the player $1$, we obtain the following belief distribution $\Delta_{0.5}(s_1)= (\{s_1 : 75\%; s'_1 : 25\%\}, \{s_2 : 50\%; s'_2:50\%\}$. $w_1$ and $w_2$ are more probable ($37.5\%$ each) than $w_3$ and $w_4$ ($12.5\%$ each). Nevertheless, their probabilities are not zero, which makes it consistent with the other player's belief.  
\\\\
It is possible to expand this concept by considering that $\lambda$ depends on the progress of the game. As an example, in trick-taking card games, it may be important to keep the private information hidden at the beginning of the game (so as not to reveal information) but, as the game progresses, the focus shifts to accumulating points before the end, where the importance of concealing this information may decrease.


\subsection{Adaptation of algorithms}

PIMC and IS-MCTS have been created with private belief distribution in mind. Therefore, it is necessary to modify the algorithms to use the public or a mixture belief distribution. Instead of starting at an infostate $s_i$, the algorithms must be adapted to start at $s_{pub}$, where $s_{pub}$ is the public infostate associated with $s_i$.

\subsubsection{PIMC}

In the case of PIMC, one must use a distinct PIMC for each infostate possible ($S_i(s_{pub})$), and combine the final result by aggregating the scores using the distribution of possible infostates ($\Delta S_i(s_{pub})$). 
%
%
\\\\
In our example, when using the mixture belief distribution, two infostates are possible for the first player ($s_1$ and $s'_1$). If $w_2$ or $w_1$ are sampled, the algorithm used is the one defined for $s_1$, on the other hand, if $w_3$ or $w_4$ are sampled, the algorithm used is the one defined for $s'_1$. In the end, if $s_1$ has been visited $75\%$ (corresponding to the mixture belief distribution with $\lambda=0.5$), the action chosen in $s_1$ will have more impact than the action chosen in $s'_1$.

\subsubsection{IS-MCTS}
\label{subsec::is_mcts}

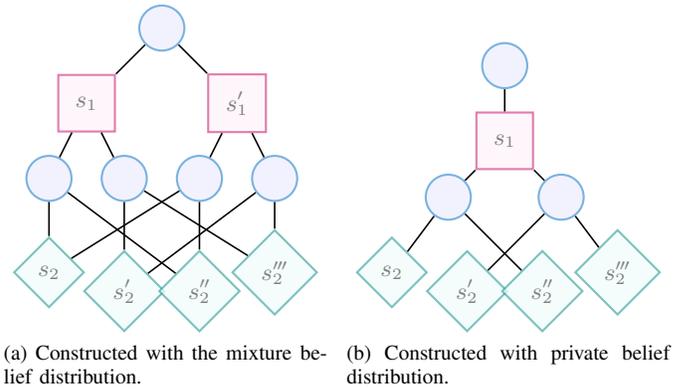
\begin{figure}[!htbp]
    \subfloat[Constructed with the mixture belief distribution.]{
    \centering

    \begin{tikzpicture}[auto, semithick, square/.style={regular polygon,regular polygon sides=4}] %
    
    \node[chance] (n0) at (0, 0) {};
    
    \node[maxnod] (n1) at (-1,-1) {\small $s_1$};
    \node[maxnod] (n2) at (1,-1) {\small $s'_1$};

    \node[chance] (n3) at (-1.5,-2) {};
    \node[chance] (n4) at (-0.5, -2) {};
    \node[chance] (n5) at (0.5,-2) {};
    \node[chance] (n6) at (1.5,-2) {};

    \node[minnod] (n7) at (-1.5,-3.25) {\small $s_2$};
    \node[minnod] (n8) at (-0.5, -3.5) {\small $s'_2$};
    \node[minnod] (n9) at (0.5,-3.5) {\small $s''_2$};
    \node[minnod] (n10) at (1.5,-3.25) {\small $s'''_2$};
    
    \draw (n0) -- (n1);
    \draw (n0) -- (n2);

    \draw (n1) -- (n3);
    \draw (n1) -- (n4);
    
    \draw (n2) -- (n5);
    \draw (n2) -- (n6);

    \draw (n3) -- (n7);
    \draw (n3) -- (n9);

    \draw (n4) -- (n8);
    \draw (n4) -- (n10);

    \draw (n5) -- (n7);
    \draw (n5) -- (n9);

    \draw (n6) -- (n8);
    \draw (n6) -- (n10);

    \end{tikzpicture}
    }
    \hfill
    \subfloat[Constructed with private belief distribution.]{
    \centering
    \begin{tikzpicture}[auto, semithick, square/.style={regular polygon,regular polygon sides=4}] %
    
    \node[chance] (n0) at (0, 0) {};
    
    \node[maxnod] (n1) at (0,-1) {\small$s_1$};

    \node[chance] (n3) at (-0.75,-1.75) {};
    \node[chance] (n4) at (0.75, -1.75) {};

    \node[minnod] (n7) at (-1.5,-2.75) {\small $s_2$};
    \node[minnod] (n8) at (-0.5, -3) {\small $s'_2$};
    \node[minnod] (n9) at (0.5,-3) {\small $s''_2$};
    \node[minnod] (n10) at (1.5,-2.75) {\small $s'''_2$};
    
    \draw (n0) -- (n1);

    \draw (n1) -- (n3);
    \draw (n1) -- (n4);

    \draw (n3) -- (n7);
    \draw (n3) -- (n9);

    \draw (n4) -- (n8);
    \draw (n4) -- (n10);    

    \end{tikzpicture}
    }
    
    \caption{Example of the tree constructed by IS-MCTS. The first player is acting in the red square, the second player is acting in the green diamond and the blue circle refers to the chance node.}
    \label{fig:is_mcts}
\end{figure}

With IS-MCTS, a singular algorithm is feasible as IS-MCTS creates a tree where the nodes represent infostates, and an infostate for player $j$ may come from several infostates of player $i$.
\\\\
An example is provided in Fig. \ref{fig:is_mcts}. For the first player, two infostates are possible ($s_1$ and $s'_1$) and four infostates are possible for the second player after the first player's action ($s_2=\{o^1_{priv(2)}= \emptyset, o^2_{priv(2)}=1, o^3_{priv(2)}=a_1\}$, $s'_2=\{o^1_{priv(2)}= \emptyset, o^2_{priv(2)}=1, o^3_{priv(2)}=a_2\}$, $s''_2=\{o^1_{priv(2)}= \emptyset, o^2_{priv(2)}=2, o^3_{priv(2)}=a_1\}$ or $s'''_2=\{o^1_{priv(2)}= \emptyset, o^2_{priv(2)}=2, o^3_{priv(2)}=a_2\}$). 

For the second player, all infostates are achievable through any infostate of the first player. For example, $s_2$ is achievable when sampling $w_1$ (from $s_1$) or when sampling $w_2$ (from $s'_1)$ and playing the action $a_1$.

\section{Experimentation}
\label{sec:exp}
\subsection{Benchmarks}

For our experiments, the following benchmarks are tested `Liar's Dice' (LD), `Card Games (CG)', and `Leduc Poker' (LP). Each of them is described below.  

\subsubsection{Card game}

For the purpose of the experimentation, we use a smaller version of classic trick-taking card games. The game is played with two players, $10/20$ cards known by all, $2/6$ are hidden and the rest is distributed to each player. 

The playing phase is decomposed into tricks, the player starting the trick is the one who won the previous trick. The starting player of a trick can play any card in his hand, but the other players must follow the suit of the first player. If they can not, they can play any card they want but, without the possibility of winning the trick. The winner of the trick is the one with the highest-ranking card. At the end of the game, the points of each player are counted (plain version of trick-taking card game). The count is defined by the number of tricks won. A player wins if it has at least half of the points. 

\subsubsection{Liar's Dice}
\label{subsub:liarDice}
Liar's dice is a dice game played with two or more players, where each player possesses $N$ dice of $K$ sides and in which a player must deceive and be able to detect an opponent's deception. 

In the beginning, each player rolls his dice and observes the values. After that, players take turns guessing the number of dice of a particular type held by everyone. The game continues until a player accuses another of lying. If the player who made the assumption is right, he wins the game, on the opposite, if the challenged player did not lie, the challenged player wins. 

During the game, a player can not bid less than previously, \ie he must at least bid more dice than the previous player’s bid, or the same number of dice but with a higher value. Lastly, the highest face is a wild card, \ie the value can be used to count for any other face.

\subsubsection{Leduc Poker}
\label{subsub:leduc_poker}

Leduc Poker, as described in \cite{southey2012bayes}, is a variation of poker that uses a deck with only two suits, each containing three cards. 

The game consists of two rounds. In the first round, each player is dealt a single private card. In the second round, a single board card is revealed. The maximum number of bets allowed is two, with the first round allowing raises of $2$ and the second round allowing raises of $4$. Both players begin the first round with $1$ already in the pot.

\subsection{Experimentation}

In our experiments, our objective is (\romannumeral 1) to observe the extent to which an algorithm $X$ reveals information according to mixture belief distribution; (\romannumeral 2) to analyze how the mixture belief distribution impacts the performance against an opponent that uses the revealed information; and (\romannumeral 3) to analyze how the mixture belief distribution impacts the performance against an opponent that does not use the revealed information. 
\\\\
Our code is based on OpenSpiel \cite{lanctot_openspiel_2019}. This is a collection of environments and algorithms for research in general reinforcement learning and search/planning in games.

PIMC and IS-MCTS are used with their basic version, \ie PIMC uses AlphaBeta and IS-MCTS uses random rollouts as the perfect information evaluator and an exploration constant of $0.7$. For both, $1000$ world states are sampled.  

To achieve a stable policy (as PIMC and IS-MCTS are online algorithms), we run the algorithm multiple times for
every infostate until the policy obtained has less than $1\%$ of variation. 

The experiments were conducted according to the player's playing position (each position reveals more or less information). In the following part, the experiments are carried out for the first player and in the appendix for the second player.

\subsubsection{How much information is revealed}

\begin{figure}[!htbp]
\centering  
	\subfloat[\centering Liar's dice with 2 dice ]{\scalebox{0.5}{
\begin{tikzpicture}

\definecolor{darkgray176}{RGB}{176,176,176}
\definecolor{darkorange25512714}{RGB}{255,127,14}
\definecolor{lightgray204}{RGB}{204,204,204}
\definecolor{steelblue31119180}{RGB}{31,119,180}

\begin{axis}[
legend cell align={left},
legend style={fill opacity=0.8, draw opacity=1, text opacity=1, draw=lightgray204},
tick align=outside,
tick pos=left,
x grid style={darkgray176},
xlabel={Lambda value},
xmin=-0.05, xmax=1.05,
xtick style={color=black},
y grid style={darkgray176},
ylabel={TSSR value},
ymin=1, ymax=10.0824399317074,
ytick style={color=black}
]
\path [draw=steelblue31119180, semithick]
(axis cs:0,9.0694071800095)
--(axis cs:0,9.6520128199905);

\path [draw=steelblue31119180, semithick]
(axis cs:0.1,8.8255471800095)
--(axis cs:0.1,9.4081528199905);

\path [draw=steelblue31119180, semithick]
(axis cs:0.2,8.7478771800095)
--(axis cs:0.2,9.3304828199905);

\path [draw=steelblue31119180, semithick]
(axis cs:0.3,8.42554132353259)
--(axis cs:0.3,9.05209867646741);

\path [draw=steelblue31119180, semithick]
(axis cs:0.4,8.23670132353259)
--(axis cs:0.4,8.86325867646741);

\path [draw=steelblue31119180, semithick]
(axis cs:0.5,7.52808657399028)
--(axis cs:0.5,8.22899342600972);

\path [draw=steelblue31119180, semithick]
(axis cs:0.6,5.81208629482618)
--(axis cs:0.6,6.56051370517382);

\path [draw=steelblue31119180, semithick]
(axis cs:0.7,3.90234926922261)
--(axis cs:0.7,4.59823073077739);

\path [draw=steelblue31119180, semithick]
(axis cs:0.8,1.74380802697205)
--(axis cs:0.8,2.04303197302795);

\path [draw=steelblue31119180, semithick]
(axis cs:0.9,1.04448917652881)
--(axis cs:0.9,1.07359082347119);

\path [draw=steelblue31119180, semithick]
(axis cs:1,1.04532039149597)
--(axis cs:1,1.07333960850403);

\path [draw=darkorange25512714, semithick]
(axis cs:0,1.23229499485969)
--(axis cs:0,1.32260500514031);

\path [draw=darkorange25512714, semithick]
(axis cs:0.1,1.19591686822186)
--(axis cs:0.1,1.27144313177814);

\path [draw=darkorange25512714, semithick]
(axis cs:0.2,1.12495796595819)
--(axis cs:0.2,1.18702203404181);

\path [draw=darkorange25512714, semithick]
(axis cs:0.3,1.09764349617825)
--(axis cs:0.3,1.14491650382175);

\path [draw=darkorange25512714, semithick]
(axis cs:0.4,1.07160668683149)
--(axis cs:0.4,1.11021331316851);

\path [draw=darkorange25512714, semithick]
(axis cs:0.5,1.07657724191886)
--(axis cs:0.5,1.11506275808114);

\path [draw=darkorange25512714, semithick]
(axis cs:0.6,1.05594190936599)
--(axis cs:0.6,1.08709809063401);

\path [draw=darkorange25512714, semithick]
(axis cs:0.7,1.0511648161887)
--(axis cs:0.7,1.0811351838113);

\path [draw=darkorange25512714, semithick]
(axis cs:0.8,1.04697908384712)
--(axis cs:0.8,1.07564091615288);

\path [draw=darkorange25512714, semithick]
(axis cs:0.9,1.04655337752043)
--(axis cs:0.9,1.07504662247957);

\path [draw=darkorange25512714, semithick]
(axis cs:1,1.04347058565312)
--(axis cs:1,1.07184941434688);

\addplot [semithick, steelblue31119180, dashed, mark=*, mark size=3, mark options={solid}]
table {%
0 9.36071
0.1 9.11685
0.2 9.03918
0.3 8.73882
0.4 8.54998
0.5 7.87854
0.6 6.1863
0.7 4.25029
0.8 1.89342
0.9 1.05904
1 1.05933
};
\addlegendentry{PIMC}
\addplot [semithick, darkorange25512714, mark=*, mark size=3, mark options={solid}]
table {%
0 1.27745
0.1 1.23368
0.2 1.15599
0.3 1.12128
0.4 1.09091
0.5 1.09582
0.6 1.07152
0.7 1.06615
0.8 1.06131
0.9 1.0608
1 1.05766
};
\addlegendentry{IS-MCTS}
\end{axis}

\end{tikzpicture}}}
 	\subfloat[\centering Liar's dice with 3 dice ]{\scalebox{0.5}{
\begin{tikzpicture}

\definecolor{darkgray176}{RGB}{176,176,176}
\definecolor{darkorange25512714}{RGB}{255,127,14}
\definecolor{lightgray204}{RGB}{204,204,204}
\definecolor{steelblue31119180}{RGB}{31,119,180}

\begin{axis}[
legend cell align={left},
legend style={fill opacity=0.8, draw opacity=1, text opacity=1, draw=lightgray204},
tick align=outside,
tick pos=left,
x grid style={darkgray176},
xlabel={Lambda value},
xmin=-0.05, xmax=1.05,
xtick style={color=black},
y grid style={darkgray176},
ylabel={TSSR value},
ymin=1, ymax=9.44243753442335,
ytick style={color=black}
]
\path [draw=steelblue31119180, semithick]
(axis cs:0,8.07972979277675)
--(axis cs:0,9.04843020722325);

\path [draw=steelblue31119180, semithick]
(axis cs:0.1,7.4818454846024)
--(axis cs:0.1,8.3613545153976);

\path [draw=steelblue31119180, semithick]
(axis cs:0.2,7.1255986753726)
--(axis cs:0.2,7.8805613246274);

\path [draw=steelblue31119180, semithick]
(axis cs:0.3,6.38232442929515)
--(axis cs:0.3,6.99649557070485);

\path [draw=steelblue31119180, semithick]
(axis cs:0.4,5.98653506539743)
--(axis cs:0.4,6.56520493460257);

\path [draw=steelblue31119180, semithick]
(axis cs:0.5,5.67467950910538)
--(axis cs:0.5,6.19638049089462);

\path [draw=steelblue31119180, semithick]
(axis cs:0.6,4.51173400841094)
--(axis cs:0.6,4.96260599158906);

\path [draw=steelblue31119180, semithick]
(axis cs:0.7,3.66926114357129)
--(axis cs:0.7,4.07245885642871);

\path [draw=steelblue31119180, semithick]
(axis cs:0.8,2.67158239205611)
--(axis cs:0.8,2.97557760794389);

\path [draw=steelblue31119180, semithick]
(axis cs:0.9,1.57840941274445)
--(axis cs:0.9,1.71745058725555);

\path [draw=steelblue31119180, semithick]
(axis cs:1,1.17681350819244)
--(axis cs:1,1.23366649180756);

\path [draw=darkorange25512714, semithick]
(axis cs:0,1.21365243051218)
--(axis cs:0,1.27488756948782);

\path [draw=darkorange25512714, semithick]
(axis cs:0.1,1.22655979430997)
--(axis cs:0.1,1.29088020569003);

\path [draw=darkorange25512714, semithick]
(axis cs:0.2,1.20510433370723)
--(axis cs:0.2,1.28545566629277);

\path [draw=darkorange25512714, semithick]
(axis cs:0.3,1.20821135724817)
--(axis cs:0.3,1.26496864275183);

\path [draw=darkorange25512714, semithick]
(axis cs:0.4,1.20011247758589)
--(axis cs:0.4,1.25616752241411);

\path [draw=darkorange25512714, semithick]
(axis cs:0.5,1.19676950683479)
--(axis cs:0.5,1.25483049316521);

\path [draw=darkorange25512714, semithick]
(axis cs:0.6,1.16980717201192)
--(axis cs:0.6,1.23963282798808);

\path [draw=darkorange25512714, semithick]
(axis cs:0.7,1.17945977446602)
--(axis cs:0.7,1.25468022553398);

\path [draw=darkorange25512714, semithick]
(axis cs:0.8,1.1774180001883)
--(axis cs:0.8,1.2409819998117);

\path [draw=darkorange25512714, semithick]
(axis cs:0.9,1.17796543780758)
--(axis cs:0.9,1.23517456219242);

\path [draw=darkorange25512714, semithick]
(axis cs:1,1.16828366322121)
--(axis cs:1,1.22287633677879);

\addplot [semithick, steelblue31119180, dashed, mark=*, mark size=3, mark options={solid}]
table {%
0 8.56408
0.1 7.9216
0.2 7.50308
0.3 6.68941
0.4 6.27587
0.5 5.93553
0.6 4.73717
0.7 3.87086
0.8 2.82358
0.9 1.64793
1 1.20524
};
\addlegendentry{PIMC}
\addplot [semithick, darkorange25512714, mark=*, mark size=3, mark options={solid}]
table {%
0 1.24427
0.1 1.25872
0.2 1.24528
0.3 1.23659
0.4 1.22814
0.5 1.2258
0.6 1.20472
0.7 1.21707
0.8 1.2092
0.9 1.20657
1 1.19558
};
\addlegendentry{IS-MCTS}
\end{axis}

\end{tikzpicture}}}
  
    \subfloat[\centering Leduc poker]{\scalebox{0.5}{
\begin{tikzpicture}

\definecolor{darkgray176}{RGB}{176,176,176}
\definecolor{darkorange25512714}{RGB}{255,127,14}
\definecolor{lightgray204}{RGB}{204,204,204}
\definecolor{steelblue31119180}{RGB}{31,119,180}

\begin{axis}[
legend cell align={left},
legend style={fill opacity=0.8, draw opacity=1, text opacity=1, draw=lightgray204},
tick align=outside,
tick pos=left,
x grid style={darkgray176},
xlabel={Lambda value},
xmin=-0.05, xmax=1.05,
xtick style={color=black},
y grid style={darkgray176},
ylabel={TSSR value},
ymin=1, ymax=2.25641616037294,
ytick style={color=black}
]
\path [draw=steelblue31119180, semithick]
(axis cs:0,2.08981089040651)
--(axis cs:0,2.19672910959349);

\path [draw=steelblue31119180, semithick]
(axis cs:0.1,1.86909240895162)
--(axis cs:0.1,1.96740759104838);

\path [draw=steelblue31119180, semithick]
(axis cs:0.2,1.6764953769643)
--(axis cs:0.2,1.7726846230357);

\path [draw=steelblue31119180, semithick]
(axis cs:0.3,1.49054491207108)
--(axis cs:0.3,1.57179508792892);

\path [draw=steelblue31119180, semithick]
(axis cs:0.4,1.31109179518153)
--(axis cs:0.4,1.38522820481847);

\path [draw=steelblue31119180, semithick]
(axis cs:0.5,1.26617273183372)
--(axis cs:0.5,1.32974726816628);

\path [draw=steelblue31119180, semithick]
(axis cs:0.6,1.1788772899566)
--(axis cs:0.6,1.2321427100434);

\path [draw=steelblue31119180, semithick]
(axis cs:0.7,1.09353292655039)
--(axis cs:0.7,1.13304707344961);

\path [draw=steelblue31119180, semithick]
(axis cs:0.8,1.03372475846999)
--(axis cs:0.8,1.05935524153001);

\path [draw=steelblue31119180, semithick]
(axis cs:0.9,1.00983706749009)
--(axis cs:0.9,1.02438293250991);

\path [draw=steelblue31119180, semithick]
(axis cs:1,1.00298809400453)
--(axis cs:1,1.01127190599547);

\path [draw=darkorange25512714, semithick]
(axis cs:0,1.26637965243346)
--(axis cs:0,1.31724034756654);

\path [draw=darkorange25512714, semithick]
(axis cs:0.1,1.26612507581681)
--(axis cs:0.1,1.31471492418319);

\path [draw=darkorange25512714, semithick]
(axis cs:0.2,1.28333183550576)
--(axis cs:0.2,1.34112816449424);

\path [draw=darkorange25512714, semithick]
(axis cs:0.3,1.17056248568555)
--(axis cs:0.3,1.21429751431445);

\path [draw=darkorange25512714, semithick]
(axis cs:0.4,1.1130696407857)
--(axis cs:0.4,1.1516503592143);

\path [draw=darkorange25512714, semithick]
(axis cs:0.5,1.10065741927913)
--(axis cs:0.5,1.14148258072087);

\path [draw=darkorange25512714, semithick]
(axis cs:0.6,1.05611766477194)
--(axis cs:0.6,1.08778233522806);

\path [draw=darkorange25512714, semithick]
(axis cs:0.7,1.03857449448341)
--(axis cs:0.7,1.06078550551659);

\path [draw=darkorange25512714, semithick]
(axis cs:0.8,1.0093916317839)
--(axis cs:0.8,1.0256683682161);

\path [draw=darkorange25512714, semithick]
(axis cs:0.9,1.00376705373638)
--(axis cs:0.9,1.01309294626362);

\path [draw=darkorange25512714, semithick]
(axis cs:1,1.00405970477223)
--(axis cs:1,1.01602029522777);

\addplot [semithick, steelblue31119180, dashed, mark=*, mark size=3, mark options={solid}]
table {%
0 2.14327
0.1 1.91825
0.2 1.72459
0.3 1.53117
0.4 1.34816
0.5 1.29796
0.6 1.20551
0.7 1.11329
0.8 1.04654
0.9 1.01711
1 1.00713
};
\addlegendentry{PIMC}
\addplot [semithick, darkorange25512714, mark=*, mark size=3, mark options={solid}]
table {%
0 1.29181
0.1 1.29042
0.2 1.31223
0.3 1.19243
0.4 1.13236
0.5 1.12107
0.6 1.07195
0.7 1.04968
0.8 1.01753
0.9 1.00843
1 1.01004
};
\addlegendentry{IS-MCTS}
\end{axis}

\end{tikzpicture}}}
    \subfloat[\centering Card Game with 10 cards]{\scalebox{0.5}{
\begin{tikzpicture}

\definecolor{darkgray176}{RGB}{176,176,176}
\definecolor{darkorange25512714}{RGB}{255,127,14}
\definecolor{lightgray204}{RGB}{204,204,204}
\definecolor{steelblue31119180}{RGB}{31,119,180}

\begin{axis}[
legend cell align={left},
legend style={fill opacity=0.8, draw opacity=1, text opacity=1, draw=lightgray204},
tick align=outside,
tick pos=left,
x grid style={darkgray176},
xlabel={Lambda value},
xmin=-0.05, xmax=1.05,
xtick style={color=black},
y grid style={darkgray176},
ylabel={TSSR value},
ymin=1, ymax=1.81963981099863,
ytick style={color=black}
]
\path [draw=steelblue31119180, semithick]
(axis cs:0,1.69358527053278)
--(axis cs:0,1.78239472946722);

\path [draw=steelblue31119180, semithick]
(axis cs:0.1,1.62445055957976)
--(axis cs:0.1,1.70962944042024);

\path [draw=steelblue31119180, semithick]
(axis cs:0.2,1.63713505952095)
--(axis cs:0.2,1.72122494047905);

\path [draw=steelblue31119180, semithick]
(axis cs:0.3,1.64716771647334)
--(axis cs:0.3,1.73089228352666);

\path [draw=steelblue31119180, semithick]
(axis cs:0.4,1.62615897020245)
--(axis cs:0.4,1.71412102979755);

\path [draw=steelblue31119180, semithick]
(axis cs:0.5,1.62077097585671)
--(axis cs:0.5,1.70698902414329);

\path [draw=steelblue31119180, semithick]
(axis cs:0.6,1.57578289019848)
--(axis cs:0.6,1.65987710980152);

\path [draw=steelblue31119180, semithick]
(axis cs:0.7,1.55483367203402)
--(axis cs:0.7,1.63244632796598);

\path [draw=steelblue31119180, semithick]
(axis cs:0.8,1.39915067816076)
--(axis cs:0.8,1.47322932183924);

\path [draw=steelblue31119180, semithick]
(axis cs:0.9,1.21985758327503)
--(axis cs:0.9,1.27692241672497);

\path [draw=steelblue31119180, semithick]
(axis cs:1,1.05191406910404)
--(axis cs:1,1.08264593089596);

\path [draw=darkorange25512714, semithick]
(axis cs:0,1.41422046803613)
--(axis cs:0,1.47365953196387);

\path [draw=darkorange25512714, semithick]
(axis cs:0.1,1.41341769475914)
--(axis cs:0.1,1.47446230524086);

\path [draw=darkorange25512714, semithick]
(axis cs:0.2,1.40405956549848)
--(axis cs:0.2,1.46288043450152);

\path [draw=darkorange25512714, semithick]
(axis cs:0.3,1.37980383394756)
--(axis cs:0.3,1.43825616605244);

\path [draw=darkorange25512714, semithick]
(axis cs:0.4,1.40700925891223)
--(axis cs:0.4,1.46507074108777);

\path [draw=darkorange25512714, semithick]
(axis cs:0.5,1.40072357254506)
--(axis cs:0.5,1.46039642745494);

\path [draw=darkorange25512714, semithick]
(axis cs:0.6,1.37242199074664)
--(axis cs:0.6,1.42977800925336);

\path [draw=darkorange25512714, semithick]
(axis cs:0.7,1.32516578093813)
--(axis cs:0.7,1.38111421906187);

\path [draw=darkorange25512714, semithick]
(axis cs:0.8,1.26139261128593)
--(axis cs:0.8,1.31334738871407);

\path [draw=darkorange25512714, semithick]
(axis cs:0.9,1.14510808910979)
--(axis cs:0.9,1.18269191089021);

\path [draw=darkorange25512714, semithick]
(axis cs:1,1.03749309883903)
--(axis cs:1,1.05920690116097);

\addplot [semithick, steelblue31119180, dashed, mark=*, mark size=3, mark options={solid}]
table {%
0 1.73799
0.1 1.66704
0.2 1.67918
0.3 1.68903
0.4 1.67014
0.5 1.66388
0.6 1.61783
0.7 1.59364
0.8 1.43619
0.9 1.24839
1 1.06728
};
\addlegendentry{PIMC}
\addplot [semithick, darkorange25512714, mark=*, mark size=3, mark options={solid}]
table {%
0 1.44394
0.1 1.44394
0.2 1.43347
0.3 1.40903
0.4 1.43604
0.5 1.43056
0.6 1.4011
0.7 1.35314
0.8 1.28737
0.9 1.1639
1 1.04835
};
\addlegendentry{IS-MCTS}
\end{axis}

\end{tikzpicture}}}

    \caption{\centering Average TSSR for IS-MCTS and PIMC on multiple benchmarks according to $\lambda$ of the mixture distribution.}

\label{fig:tssr}
\end{figure}
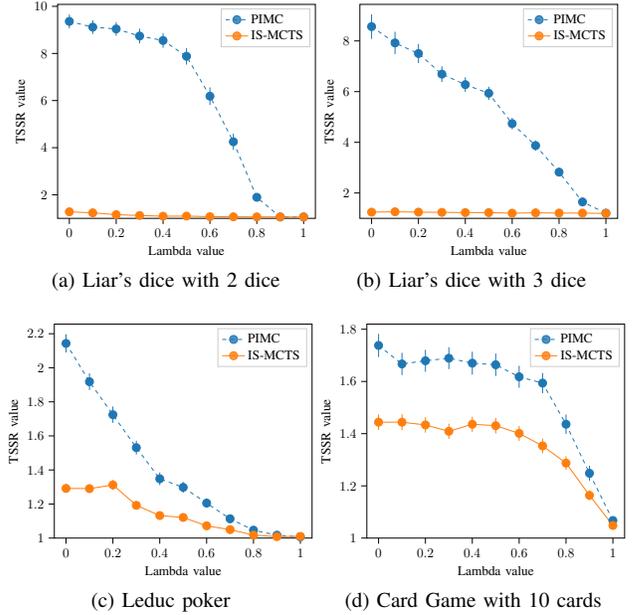

For analyzing the impact of the revealed information according to the distribution used, we use the formula called True State Sampling Ratio (TSSR) \cite{solinas_improving_2019}. TSSR measures how much more likely it is for the opponent to guess the current world state when using an algorithm $X$ rather than using a uniform function. 

The formula is $TSSR(w) = \eta(w \mid s_i ) \cdot |S_i(s_i)|$ where $s_i$ is the infostate corresponding to $w$, $\eta(w \mid s_i)$ is the probability that the true state is guessed given the information set $s_i$. The closer the result is to $1$, the less likely it is to know the real world state. Fig. \ref{fig:tssr} presents the TSSR value obtained according to $\lambda$ of the mixture distribution. 
\\\\
As expected, playing closer to the public belief distribution greatly reduces the probability of knowing the real-world state. In `Liar's Dice' with $2$ dice with PIMC, it is up to $10$ fold more likely to guess the real world state when using the private instead of the public belief distribution.

In terms of information revealed, we observe that PIMC reveals more information than IS-MCTS in every benchmark. In `Leduc Poker', it's up to $2.2$ times more likely to deduce the true state with PIMC at $\lambda =0.0$ whereas, with IS-MCTS, it is `only' $1.3$ times more likely to deduce the true state. 

In addition, `Liar's Dice' is the game that reveals the most information with the algorithm revealing up to $10$ times more likely than random, whereas in `Leduc Poker' or `Card Game', it is only up to $2$ times more likely than random.

For the following experiments, it is expected to observe $\lambda$ closer to 1 for PIMC in `Liar's Dice', as it reveals more information, and therefore, could be exploited by the opponent.

\subsubsection{How does the mixture impact the performance}

\begin{table*}[htbp]
\captionof{table}{\centering Expected utility against best responder when playing at the first player position.}
\centering
\begin{tabular}{*{13}{c} }
\toprule

\multirow{2}{*}{Algo} & \multirow{2}{*}{Game} & \multicolumn{11}{c}{$\lambda$} \\

\cmidrule(lr){3-13}

& & 0.0 & 0.1 & 0.2 & 0.3 & 0.4 & 0.5 & 0.6 & 0.7 & 0.8 & 0.9 & 1 \\ 

\midrule



\multirow{3}{*}{PIMC} & LD 2D & 0.300 & 0.298 & 0.297 & 0.292 & 0.294 & 0.288 & \textbf{0.281} & 0.290 & 0.336 & 0.382 & 0.382 \\
& LD 3D & 0.313 & 0.276 & 0.265 & 0.269 & \textbf{0.235} & 0.283 & 0.324 & 0.356 & 0.359 & 0.393 & 0.458  \\
& LP & 0.622 & \textbf{0.616} & 0.660 & 0.767 & 0.797 & 1.481 & 1.626 & 1.480 & 1.532 & 1.599 & 1.611 \\

\midrule

\multirow{2}{*}{IS-MCTS} & LD 2D & 0.513 & \textbf{0.512} & 0.517 & 0.528 & 0.539 & 0.547 & 0.552 & 0.554 & 0.555 & 0.562 & 0.562 \\
& LP & \textbf{0.797} & 0.890 & 0.966 & 0.959 & 1.158 & 1.226 & 1.402 & 1.673 & 1.786 & 2.083 & 2.326  \\

\bottomrule
\end{tabular} 
\label{tab:exploi_P0}
\end{table*} 

To measure how the mixture impacts the performance, we compute the expected utility against the best responder. The best responder is the worst possible enemy of all algorithms, \ie it knows exactly the policy our algorithm will execute, and therefore, can infer the true infostate and plays the best action against it. 

The results are available in Table \ref{tab:exploi_P0} where the values represent the expected utility of the best responder and must be minimized. The results obtained are exact utility (without variation), as the best responder computes the best strategy knowing all the distributions in every infostate of the game.
\\\\
We observe that the private belief distribution performs better than the public belief distribution, \ie for all benchmarks and algorithms (better results are obtained when $\lambda=0.0$ than when $\lambda=1.0$).

In `Liar's Dice' with PIMC, the best performances are obtained when $\lambda$ is close to $0.5$ (with $2$ dice, we obtain the best value when $\lambda=0.6$). These results were expected, as PIMC reveals a lot of information with Liar's Dice which is then exploited by the best responder.

On the other hand, when the algorithm reveals less information (as observed in `Leduc Poker' or IS-MCTS), it is preferable to use the private belief distribution or very close, as it is not sufficient for the best responder to exploit the revealed information.

\subsubsection{Can the use of multiple mixture belief distributions throughout the game improve performance}

In this experiment, we analyze the use of multiple mixtures throughout the game to improve performance. For this purpose, we compute multiple mixture distributions against the best responder. 

\begin{figure}[!htbp]
\centering

	\subfloat[\centering Leduc Poker with PIMC ]{\includegraphics[scale=0.33]{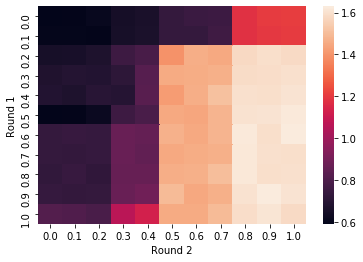}}
    \subfloat[\centering Liar's Dice 2 dice with PIMC ]{\includegraphics[scale=0.33]{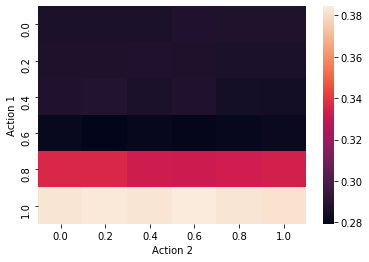}}
    
	\subfloat[\centering Leduc Poker with IS MCTS ]{\includegraphics[scale=0.33]{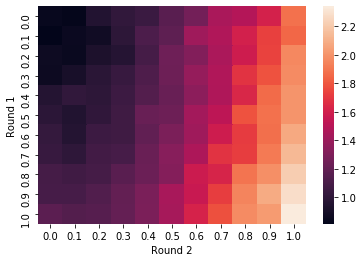}}
    \subfloat[\centering Liar's Dice 2 dice with IS MCTS ]{\includegraphics[scale=0.33]{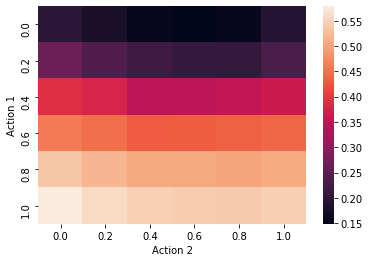}}

    \caption{\centering Heatmap of the expected utility against the best response when playing at the first position.}
\label{fig:heatmap}
\end{figure}

Fig. \ref{fig:heatmap} represents heatmaps for `Leduc Poker' and `Liar's Dice' according to the position throughout the game when using PIMC (resp. IS-MCTS). For both games, we have a mixture distribution for the first action and another for the second action. 
\\\\
In all experiments, we observe that using multiple mixtures throughout the game has an impact on the performance. 
In `Leduc Poker' for both algorithms, not using our private belief distribution is more punished in the second round than in the first round (\eg $\{0.0, 1.0\}$ has a value of $1.17$ whereas $\{1.0, 0.0\}$ has a value of $1.88$ for IS-MCTS). On the other hand, for `Liar's Dice', we observe that the first round is the most important one. 

In addition, we observe that playing multiple mixtures improve performance. In `Liar's Dice', the best value for IS MCTS is obtained when we have $\{0.0, 0.6\}$ and for PIMC when we have $\{0.6, 0.2\}$.

\subsubsection{How does the mixture impact the winning rate}

\begin{table*}[htbp]
\captionof{table}{\centering Winning rate when the opponent uses `PIMC' when playing at the first player position. } 
\centering
\begin{tabular}{*{13}{c} }
\toprule

\multirow{2}{*}{Our} & \multirow{2}{*}{Game} & \multicolumn{11}{c}{$\lambda$} \\

\cmidrule(lr){3-13}

&  & 0.0 & 0.1 & 0.2 & 0.3 & 0.4 & 0.5 & 0.6 & 0.7 & 0.8 & 0.9 & 1 \\ 

\midrule

\multirow{4}{*}{PIMC} & LD 3D & 48.6 & \textbf{50.4}  & 47.9 & 47.4 & 44.6 & 42.6 & 39.9 & 37.5 & 36.1 & 28.9 & 27.8\\
& LD 5D & 43.1 & \textbf{43.4} & 42.2 & \textbf{43.4} & 42.5 & 41.4 & 39.8 & 36.3 & 37.1 & 29.8 & 23.6 \\
& CG 10C & \textbf{48.2} & 47.9 & 47.7 & 47.7 & 47.6 & 47.4 & 46.7 & 46. & 45.5 & 39.8 & 31.6 \\
& CG 20C & 53.7 & 53.8 & 54.2 & \textbf{54.5} & 53.9 & 53.2 & 52.8 & 52. & 47.4 & 36.3 & 23.5 \\

\cmidrule(lr){1-13}

\multirow{4}{*}{IS MCTS} & LD 3D & 23.7 & 23.7 & 24.7 & \textbf{27.} & 23.1 & 23.1 & 21.7 & 20. & 19.3 & 15.4 & 16.4\\
& LD 5D & 22. & 20.9  & 21.9 & \textbf{22.2} & 21.9 & 20.8 & 21.6 & 21. & 16.9 & 15.5 & 13.4 \\
& CG 10C & 45.3 & \textbf{46.3} & 45.4 & 45.1 & 43.8 & 45.1 & 45. &  43.1 & 42.7 & 37.6 & 30.  \\
& CG 20C &  36.5 & \textbf{38.5} & 38.2 & 36.2 & 36.4 & 36.6 & 35.5 &  34.9 & 33.3 & 33.1 & 20.8\\

\bottomrule
\end{tabular}
\label{tab:winning_rate}
\end{table*}

As observed in the previous experiments, when using a $\lambda$ closer to the public belief distribution, we obtain a distribution of action less relevant but with the advantage of disclosing less information. Therefore, when faced with an opponent who does not infer on our private information, it is expected to lose the benefit of using a $\lambda$ closer to the public belief distribution. Nevertheless, using a $\lambda$ closer to the public belief distribution not only reveals less information but allows it to be more consistent with the other player’s doubts. 

To measure the impact of being more consistent with the other player's doubts, we evaluate the performance against an algorithm that does not try to infer our private information. To do this, we compute the winning rate against `PIMC' over $1000$ games which results in $3.1\%$ variation ($95\%$ of confidence interval). The scores are available in Table \ref{tab:winning_rate}.
\\\\
As before, we observe that it is preferable to use private belief distribution instead of public belief distribution. In `Liar's Dice' with 3 dice with PIMC, we observe a drop of $20.8$ in the winning rate between the private and public belief distribution. In addition, we observe that in every benchmark tested and for both algorithms, using a $\lambda$ between $0.0$ to $0.5$ does not produce a drop in performance, but provides equivalent results. 
\\\\
These results are surprising, as we could have expected a drop in performance as the actions are less relevant to the current infostate (as we have sampled less often the true infostate). This implies that being more consistent with the doubts of the other players compensates for the loss of the player’s private information.

\section{Conclusion}

In this paper, we study the strengths and weaknesses of probability distributions (private and public) in which particular attention has been paid to the revealed information and the impact of this revealed information on performance. Our study has been carried out on determinization-based algorithms and on multiple imperfect information games.


We complete the study by proposing a new probability distribution, a mixture of the two previous ones, which solves problems encountered by other distributions. We show that using the mixture is beneficial to reduce the revealed information and improve performance. We also show that using multiple mixtures throughout the game improves performance. In addition, we observed that using the mixture against an opponent that does not use our private information revealed results in a good performance as we are being more consistent with the other player's doubt.

An avenue for improvement would be to extend the utilization of using multiple mixtures throughout the game. For example, by using the mixture at each public infostate instead of a fixed time step or using a different lambda for the opponent player. Another area for improvement would be to extend the study of algorithms that do not use determinization or even, without probability distributions but bearing in mind that one should not always use one's private information at the risk of revealing information and, on the contrary, that one should not always use one's public information in order to be more consistent to one's private knowledge. Lastly, it would be interesting to extend the results at a larger scale, either by using more games or by using larger games.

\appendix

\subsection{Complementary experiments}

The following experiments are identical to those in the primary paper, with the exception that they are conducted for the second player position.

Similar results are observed, \ie PIMC reveals more information than IS-MCTS, the private belief distribution obtains better performance than the public belief distribution against the best responder, using multiple mixtures is useful to improve the performance and it is all as well to play the mixture as the private against an opponent that does not try to infer.

Yet, we also observe some differences, especially that less information is revealed when playing in the second position, which results in $\lambda$ closer to the private belief distribution against the best responder.

\begin{table*}[htbp]
\captionof{table}{\centering Expected utility for best responder against our algorithm being the second player.} 
\centering
\begin{tabular}{*{13}{c} }
\toprule

\multirow{2}{*}{Algo} & \multirow{2}{*}{Game} & \multicolumn{11}{c}{$\lambda$} \\

\cmidrule(lr){3-13}

&  & 0.0 & 0.1 & 0.2 & 0.3 & 0.4 & 0.5 & 0.6 & 0.7 & 0.8 & 0.9 & 1 \\ 

\midrule

\multirow{2}{*}{PIMC} & LD 2D & \textbf{0.678} & 0.695 & 0.703 & 0.707 & 0.716 & 0.718 & 0.711 & 0.741 & 0.779 & 0.836 & 0.836  \\
& LP & \textbf{0.398} & 0.400 & 0.459 & 0.612 & 0.796 & 1.461 & 1.450 & 1.509 & 1.593 & 1.615 & 1.632 \\

\midrule

\multirow{2}{*}{IS-MCTS} & LD 2D & 0.697 & \textbf{0.687} & 0.697 & 0.716 & 0.727 & 0.732 & 0.740 & 0.751 & 0.759 & 0.768 & 0.787 \\
& LP & \textbf{0.784} & 0.784 & 0.898 & 0.800 & 1.017 & 1.078 & 1.186 & 1.324 & 1.561 & 1.728 & 2.002 \\

\bottomrule
\end{tabular}
\label{tab:exploit_P2}
\end{table*}

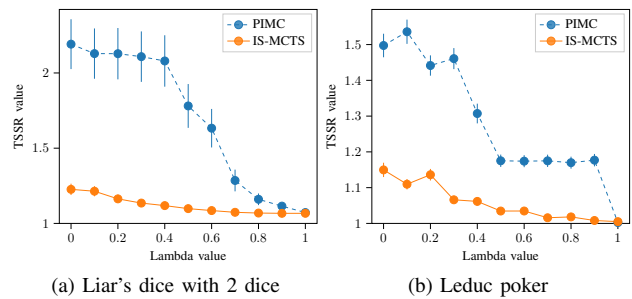
\begin{figure}[!htbp]
\centering
	
	\subfloat[\centering Liar's dice with 2 dice ]{\scalebox{0.5}{
\begin{tikzpicture}

\definecolor{darkgray176}{RGB}{176,176,176}
\definecolor{darkorange25512714}{RGB}{255,127,14}
\definecolor{lightgray204}{RGB}{204,204,204}
\definecolor{steelblue31119180}{RGB}{31,119,180}

\begin{axis}[
legend cell align={left},
legend style={fill opacity=0.8, draw opacity=1, text opacity=1, draw=lightgray204},
tick align=outside,
tick pos=left,
x grid style={darkgray176},
xlabel={Lambda value},
xmin=-0.05, xmax=1.05,
xtick style={color=black},
y grid style={darkgray176},
ylabel={TSSR value},
ymin=1, ymax=2.4221886627579,
ytick style={color=black}
]
\path [draw=steelblue31119180, semithick]
(axis cs:0,2.0258125995579)
--(axis cs:0,2.3570074004421);

\path [draw=steelblue31119180, semithick]
(axis cs:0.1,1.96162089169613)
--(axis cs:0.1,2.29583910830387);

\path [draw=steelblue31119180, semithick]
(axis cs:0.2,1.95757340715746)
--(axis cs:0.2,2.29880659284254);

\path [draw=steelblue31119180, semithick]
(axis cs:0.3,1.94075447172423)
--(axis cs:0.3,2.27612552827577);

\path [draw=steelblue31119180, semithick]
(axis cs:0.4,1.90890858362101)
--(axis cs:0.4,2.25063141637899);

\path [draw=steelblue31119180, semithick]
(axis cs:0.5,1.63515329916435)
--(axis cs:0.5,1.92638670083565);

\path [draw=steelblue31119180, semithick]
(axis cs:0.6,1.50492051069151)
--(axis cs:0.6,1.76087948930849);

\path [draw=steelblue31119180, semithick]
(axis cs:0.7,1.21319647506638)
--(axis cs:0.7,1.35750352493362);

\path [draw=steelblue31119180, semithick]
(axis cs:0.8,1.1216595547182)
--(axis cs:0.8,1.1990204452818);

\path [draw=steelblue31119180, semithick]
(axis cs:0.9,1.09612771859514)
--(axis cs:0.9,1.13377228140486);

\path [draw=steelblue31119180, semithick]
(axis cs:1,1.05821667265744)
--(axis cs:1,1.08624332734256);

\path [draw=darkorange25512714, semithick]
(axis cs:0,1.18872808124127)
--(axis cs:0,1.26399191875873);

\path [draw=darkorange25512714, semithick]
(axis cs:0.1,1.18030893306057)
--(axis cs:0.1,1.24763106693943);

\path [draw=darkorange25512714, semithick]
(axis cs:0.2,1.13571135840527)
--(axis cs:0.2,1.19070864159473);

\path [draw=darkorange25512714, semithick]
(axis cs:0.3,1.11081600963405)
--(axis cs:0.3,1.15958399036595);

\path [draw=darkorange25512714, semithick]
(axis cs:0.4,1.09766549045465)
--(axis cs:0.4,1.13907450954535);

\path [draw=darkorange25512714, semithick]
(axis cs:0.5,1.07952476327173)
--(axis cs:0.5,1.11713523672827);

\path [draw=darkorange25512714, semithick]
(axis cs:0.6,1.06972848479088)
--(axis cs:0.6,1.10111151520912);

\path [draw=darkorange25512714, semithick]
(axis cs:0.7,1.05969442571453)
--(axis cs:0.7,1.08794557428547);

\path [draw=darkorange25512714, semithick]
(axis cs:0.8,1.05505701384025)
--(axis cs:0.8,1.08264298615975);

\path [draw=darkorange25512714, semithick]
(axis cs:0.9,1.05338215412593)
--(axis cs:0.9,1.08077784587407);

\path [draw=darkorange25512714, semithick]
(axis cs:1,1.0534983058493)
--(axis cs:1,1.0806616941507);

\addplot [semithick, steelblue31119180, dashed, mark=*, mark size=3, mark options={solid}]
table {%
0 2.19141
0.1 2.12873
0.2 2.12819
0.3 2.10844
0.4 2.07977
0.5 1.78077
0.6 1.6329
0.7 1.28535
0.8 1.16034
0.9 1.11495
1 1.07223
};
\addlegendentry{PIMC}
\addplot [semithick, darkorange25512714, mark=*, mark size=3, mark options={solid}]
table {%
0 1.22636
0.1 1.21397
0.2 1.16321
0.3 1.1352
0.4 1.11837
0.5 1.09833
0.6 1.08542
0.7 1.07382
0.8 1.06885
0.9 1.06708
1 1.06708
};
\addlegendentry{IS-MCTS}
\end{axis}

\end{tikzpicture}}}
    \subfloat[\centering Leduc poker]{\scalebox{0.5}{
\begin{tikzpicture}

\definecolor{darkgray176}{RGB}{176,176,176}
\definecolor{darkorange25512714}{RGB}{255,127,14}
\definecolor{lightgray204}{RGB}{204,204,204}
\definecolor{steelblue31119180}{RGB}{31,119,180}

\begin{axis}[
legend cell align={left},
legend style={fill opacity=0.8, draw opacity=1, text opacity=1, draw=lightgray204},
tick align=outside,
tick pos=left,
x grid style={darkgray176},
xlabel={Lambda value},
xmin=-0.05, xmax=1.05,
xtick style={color=black},
y grid style={darkgray176},
ylabel={TSSR value},
ymin=1, ymax=1.59818911612943,
ytick style={color=black}
]
\path [draw=steelblue31119180, semithick]
(axis cs:0,1.46449315437814)
--(axis cs:0,1.53028684562186);

\path [draw=steelblue31119180, semithick]
(axis cs:0.1,1.501963054484)
--(axis cs:0.1,1.569716945516);

\path [draw=steelblue31119180, semithick]
(axis cs:0.2,1.41302629595213)
--(axis cs:0.2,1.46981370404787);

\path [draw=steelblue31119180, semithick]
(axis cs:0.3,1.43134435912454)
--(axis cs:0.3,1.49017564087546);

\path [draw=steelblue31119180, semithick]
(axis cs:0.4,1.27969001868617)
--(axis cs:0.4,1.33482998131384);

\path [draw=steelblue31119180, semithick]
(axis cs:0.5,1.15785677966751)
--(axis cs:0.5,1.19238322033249);

\path [draw=steelblue31119180, semithick]
(axis cs:0.6,1.15724219094013)
--(axis cs:0.6,1.19077780905987);

\path [draw=steelblue31119180, semithick]
(axis cs:0.7,1.15778541335497)
--(axis cs:0.7,1.19219458664503);

\path [draw=steelblue31119180, semithick]
(axis cs:0.8,1.15305247972672)
--(axis cs:0.8,1.18656752027328);

\path [draw=steelblue31119180, semithick]
(axis cs:0.9,1.15952663869722)
--(axis cs:0.9,1.19421336130278);

\path [draw=steelblue31119180, semithick]
(axis cs:1,1.00027353324754)
--(axis cs:1,1.00444646675246);

\path [draw=darkorange25512714, semithick]
(axis cs:0,1.12964888764567)
--(axis cs:0,1.16963111235433);

\path [draw=darkorange25512714, semithick]
(axis cs:0.1,1.09509436041693)
--(axis cs:0.1,1.12410563958307);

\path [draw=darkorange25512714, semithick]
(axis cs:0.2,1.1199712338464)
--(axis cs:0.2,1.15230876615359);

\path [draw=darkorange25512714, semithick]
(axis cs:0.3,1.05509281279084)
--(axis cs:0.3,1.07694718720916);

\path [draw=darkorange25512714, semithick]
(axis cs:0.4,1.05066840942684)
--(axis cs:0.4,1.07241159057316);

\path [draw=darkorange25512714, semithick]
(axis cs:0.5,1.02601754048297)
--(axis cs:0.5,1.04318245951703);

\path [draw=darkorange25512714, semithick]
(axis cs:0.6,1.02663480980565)
--(axis cs:0.6,1.04274519019435);

\path [draw=darkorange25512714, semithick]
(axis cs:0.7,1.00979369654157)
--(axis cs:0.7,1.02184630345843);

\path [draw=darkorange25512714, semithick]
(axis cs:0.8,1.01249588936038)
--(axis cs:0.8,1.02394411063962);

\path [draw=darkorange25512714, semithick]
(axis cs:0.9,1.00411472802369)
--(axis cs:0.9,1.01220527197631);

\path [draw=darkorange25512714, semithick]
(axis cs:1,1.00093472802369)
--(axis cs:1,1.00902527197631);

\addplot [semithick, steelblue31119180, dashed, mark=*, mark size=3, mark options={solid}]
table {%
0 1.49739
0.1 1.53584
0.2 1.44142
0.3 1.46076
0.4 1.30726
0.5 1.17512
0.6 1.17401
0.7 1.17499
0.8 1.16981
0.9 1.17687
1 1.00236
};
\addlegendentry{PIMC}
\addplot [semithick, darkorange25512714, mark=*, mark size=3, mark options={solid}]
table {%
0 1.14964
0.1 1.1096
0.2 1.13614
0.3 1.06602
0.4 1.06154
0.5 1.0346
0.6 1.03469
0.7 1.01582
0.8 1.01822
0.9 1.00816
1 1.00498
};
\addlegendentry{IS-MCTS}
\end{axis}

\end{tikzpicture}}}

    \caption{\centering Average TSSR according to $\lambda$ value of the mixture distribution.}

\label{fig:tssr_p2}
\end{figure}

\begin{figure}[!htbp]
\centering

    \subfloat[\centering Leduc poker with PIMC]{\includegraphics[scale=0.25]{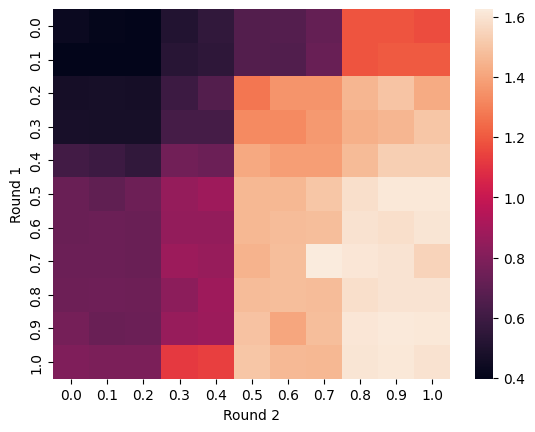}}
    \subfloat[\centering Liar's dice 2 dice with PIMC ]{\includegraphics[scale=0.25]{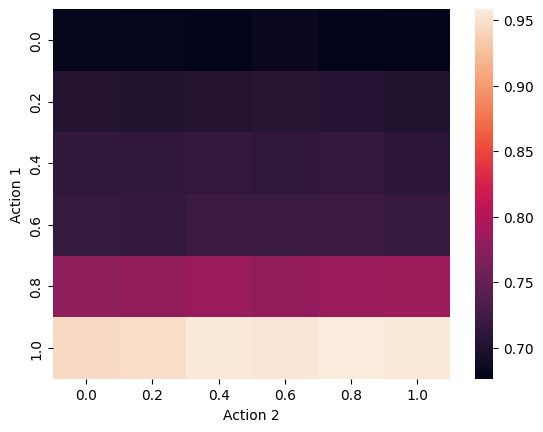}}
    
    \subfloat[\centering Leduc poker with IS-MCTS]{\includegraphics[scale=0.25]{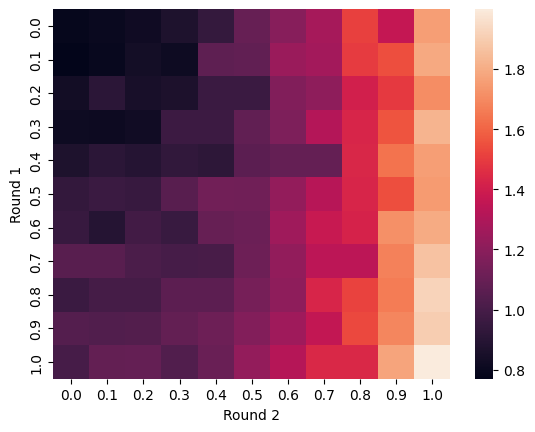}}
	\subfloat[\centering Liar's dice 2 dice with IS-MCTS]{\includegraphics[scale=0.25]{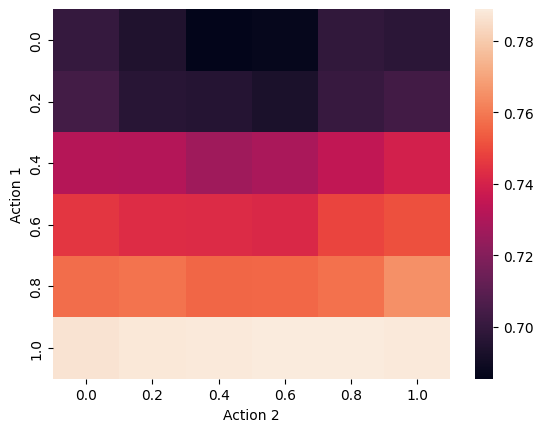}}
 
    \caption{\centering Heatmap of expected utility against the best response when playing at the second position. }
\label{fig:heatmap_PIMC_p2}
\end{figure}

\begin{table*}[htbp]
\captionof{table}{\centering Winning rate when the opponent uses `PIMC' when playing at the second player position. } 
\centering
\begin{tabular}{*{13}{c} }
\toprule

\multirow{2}{*}{Our} & \multirow{2}{*}{Game} & \multicolumn{11}{c}{$\lambda$} \\

\cmidrule(lr){3-13}

&  & 0.0 & 0.1 & 0.2 & 0.3 & 0.4 & 0.5 & 0.6 & 0.7 & 0.8 & 0.9 & 1 \\ 

\midrule

\multirow{2}{*}{PIMC} & LD 3D & 51.9 &  \textbf{53.} & 51.3 & 49.8 & 49.7 & 51.3 & 48.7 & 48.6 & 46.9 & 46.1 & 42.7 \\
& LD 5D & \textbf{56.7} & 55.5 & 56. & 56.2 & 54.8 & 56.1 & 55.3 & 53. & 51.9 & 44.7 & 42.3 \\

\cmidrule(lr){1-13}

\multirow{2}{*}{IS MCTS} & LD 3D & 48.4 & \textbf{51.3} & 49.9 & 49. & 50.1 & 51. & 47.4 & 44. & 39.7 & 36.9 & 33.3 \\
& LD 5D & \textbf{48.4} & 47.1 & 48. & 46.7 & 47.8 & 45. & 46.5 & 40.7 & 34.4 & 23.2 & 14.7\\

\bottomrule
\end{tabular}
\label{tab:winning_rate_p2}
\end{table*}

\bibliographystyle{IEEEtran}
\bibliography{IEEEabrv,Doctorat}
\end{document}